\documentclass[journal=jctc,manuscript=article]{achemso}

\usepackage{amsmath}
\usepackage{graphicx}
\usepackage{placeins}
\usepackage{hyperref}
\hypersetup{hidelinks}

\author{Suyang Zhuang}
\affiliation{State Key Laboratory of Heavy Oil Processing, China University of Petroleum (Beijing), Beijing 102249, China}
\affiliation{Zhonghai Energy Storage Technology (Beijing) Co., Ltd., Beijing 102249, China}
\author{Zekun Jiang}
\affiliation{State Key Laboratory of Heavy Oil Processing, China University of Petroleum (Beijing), Beijing 102249, China}
\affiliation{Zhonghai Energy Storage Technology (Beijing) Co., Ltd., Beijing 102249, China}
\author{Tianhang Zhou}
\affiliation{State Key Laboratory of Heavy Oil Processing, China University of Petroleum (Beijing), Beijing 102249, China}
\affiliation{Zhonghai Energy Storage Technology (Beijing) Co., Ltd., Beijing 102249, China}
\email{zhouth@cup.edu.cn}

\title{Early Cycle Charge Trajectory Generative Prediction and Full Life Cycle Health Management of Iron-Chromium Flow Batteries Based on FlowBD-E1}
\abbreviations{Fe-Cr, RFB, V/I, SOH, MAPE, RMSE, SLE, RLF, TFU}
\keywords{Iron-chromium redox flow battery, Generative forecasting, Early cycle prediction, Charge trajectory prediction, Industrial validation}

\begin{document}
\maketitle

\section*{Summary}

Long-duration stationary energy storage requires batteries whose degradation can be detected before substantial capacity loss has accumulated. Iron-chromium redox flow batteries are attractive for this role because they use abundant and low-cost active species, yet their operation is shaped by slow chromium kinetics, hydrogen evolution, membrane crossover and electrolyte imbalance. These coupled processes gradually reshape the full charge voltage/current (V/I) trajectory, but most battery prognostic studies either focus on lithium-ion cells or compress ageing into scalar capacity and state-of-health (SOH) labels. Here we study an industrial 33 kW Fe-Cr redox flow battery and introduce FlowBD-E1, an early-cycle generative forecasting framework that predicts complete future charge V/I trajectories from only the first few cycles. The model combines a multi-scale convolutional encoder, a lifecycle Transformer and an age-aware FiLM decoder, and we compare three deployment strategies: single-step latent extrapolation (SLE), recursive latent forecasting (RLF) and teacher-forced updating (TFU). Using the first 9 of 289 cycles, RLF achieved a joint V/I mean absolute percentage error (MAPE) of 0.731\% over the remaining lifecycle and produced SOH estimates below 1\% MAPE. Ablation and independent-sequence tests showed that the age-aware generative architecture outperformed LSTM and TCN baselines and retained sub-percent errors under industrial validation. These results suggest that early-cycle trajectory generation can turn a short commissioning record into a long-horizon diagnostic signal for flow-battery management. The source code is publicly available at \href{https://github.com/cupzhoutianhang/cyclegpt-battery-cycle-analysis}{github.com/cupzhoutianhang/cyclegpt-battery-cycle-analysis}.

\noindent\textit{Contributors:} Suyang Zhuang, Zekun Jiang, and Tianhang Zhou. Suyang Zhuang and Zekun Jiang contributed equally. Correspondence: \href{mailto:zhouth@cup.edu.cn}{zhouth@cup.edu.cn}.

\section{Introduction}\label{introduction}

Grid decarbonization increases the need for electrochemical storage that can operate for many hours, tolerate deep cycling and be scaled without excessive material cost \cite{ref1,ref2,ref3,ref4}. Redox flow batteries (RFBs) are particularly well matched to this requirement because power and energy are partly decoupled: the stack determines power, whereas external electrolyte tanks determine stored energy \cite{ref2,ref3,ref4}. Vanadium RFBs have become the most mature flow-battery chemistry \cite{ref5}, but the cost and supply constraints of vanadium motivate renewed interest in lower-cost aqueous chemistries. The iron-chromium (Fe-Cr) system, first demonstrated in the NASA flow-battery programme \cite{ref6}, remains especially attractive because iron and chromium are abundant, inexpensive and compatible with large stationary installations \cite{ref7,ref8}.

The same chemistry that makes Fe-Cr RFBs attractive also makes their long-term operation difficult to predict. Chromium redox kinetics are slow on conventional carbon electrodes, hydrogen evolution competes with the desired negative-electrode reaction, and cross-membrane transport gradually changes electrolyte composition and charge balance \cite{ref7,ref8,ref9,ref10,ref11,ref12}. Membrane selectivity, crossover and electrolyte lifetime are therefore central design and operational issues across flow-battery chemistries \cite{ref13,ref14,ref15}. Existing diagnostic studies have shown that RFB state variables can be inferred from electrochemical signals or model-based observers \cite{ref16,ref17}, and recent machine-learning studies have improved online state-of-charge estimation for vanadium systems \cite{ref18,ref19,ref20}. These studies are important for real-time control, but they largely estimate the present state within a cycle; they do not ask whether the first few cycles of an industrial Fe-Cr stack contain enough information to forecast how complete charge trajectories will evolve over the rest of life.

Early-cycle lifetime prediction has advanced rapidly in lithium-ion batteries. Severson et al. showed that features extracted before obvious capacity degradation can predict cycle life \cite{ref21}, and subsequent work used closed-loop optimization, engineered health features and deep learning to improve prediction across operating protocols and ageing conditions \cite{ref22,ref23,ref24,ref25,ref26,ref27}. More recently, generative and Transformer-based methods have moved from scalar lifetime prediction toward synthesis of future degradation curves or charging trajectories \cite{ref28,ref29,ref30,ref31}. However, this literature has two limitations for flow-battery management. First, it is dominated by lithium-ion cells, where the relevant electrochemical signatures and degradation modes differ from circulating-electrolyte systems. Second, many models still reduce ageing to capacity, SOH or remaining useful life; this discards the shape of voltage and current curves, even though those curves contain information about polarization, electrolyte imbalance, accessible capacity and charge termination.

For Fe-Cr RFBs, the full charge V/I trajectory is a natural health signature. Under constant-power charging, stack voltage rises and current falls as internal resistance, electrolyte state and accessible capacity change. The early voltage rise reflects ohmic and activation polarization; the mid-cycle region carries information on quasi-steady electrolyte transport and concentration state; and the terminal rise is sensitive to capacity limitation and side reactions. A model that predicts only a scalar SOH value cannot be interrogated for these phase-dependent effects. In contrast, a generative trajectory model can reconstruct the operating waveform first and allow any derived metric, including SOH, charge duration and efficiency-related quantities, to be computed afterwards.

Here we develop FlowBD-E1 for early-cycle prediction of industrial Fe-Cr RFB charge trajectories. Our central argument is that a small number of commissioning cycles can be encoded into a lifecycle representation from which future V/I waveforms and health indicators are generated with sub-percent error. The study makes three contributions. First, it treats future charge curves as the prediction target rather than as features used only to infer SOH. Second, it compares SLE, RLF and TFU strategies to separate autonomous forecasting error from diagnostic upper-bound behaviour. Third, it evaluates the method on data from a 33 kW industrial single-stack Fe-Cr system, linking recent progress in sequence modelling \cite{ref32,ref33,ref34,ref35,ref36,ref37,ref38} with a deployable flow-battery prognostic problem.

\section{Data and method}\label{data-and-method}

The data comes from long-term constant power charge and discharge operation of a 33 kW single-stack iron-chromium flow battery at the demonstration site of Zhonghai Energy Storage Technology (Beijing) Co., Ltd. Each raw record contains the cycle index, operation mode, timestamp, stack terminal voltage and stack current sampled at 1 Hz. After data cleaning, normalization and cycle segmentation, the dataset was organized into 24 independent groups, each containing 289 consecutive charge cycles. Twenty groups were used for training, three for validation and one independent industrial sequence for testing. The voltage span was 78.13-105.60 V, the current span was 312.47-428.62 A and the capacity-derived SOH decreased from 1.000 to 0.608 over the 289-cycle sequence.

\begin{figure*}[!t]
\centering
\includegraphics[width=0.70\textwidth,height=0.76\textheight,keepaspectratio]{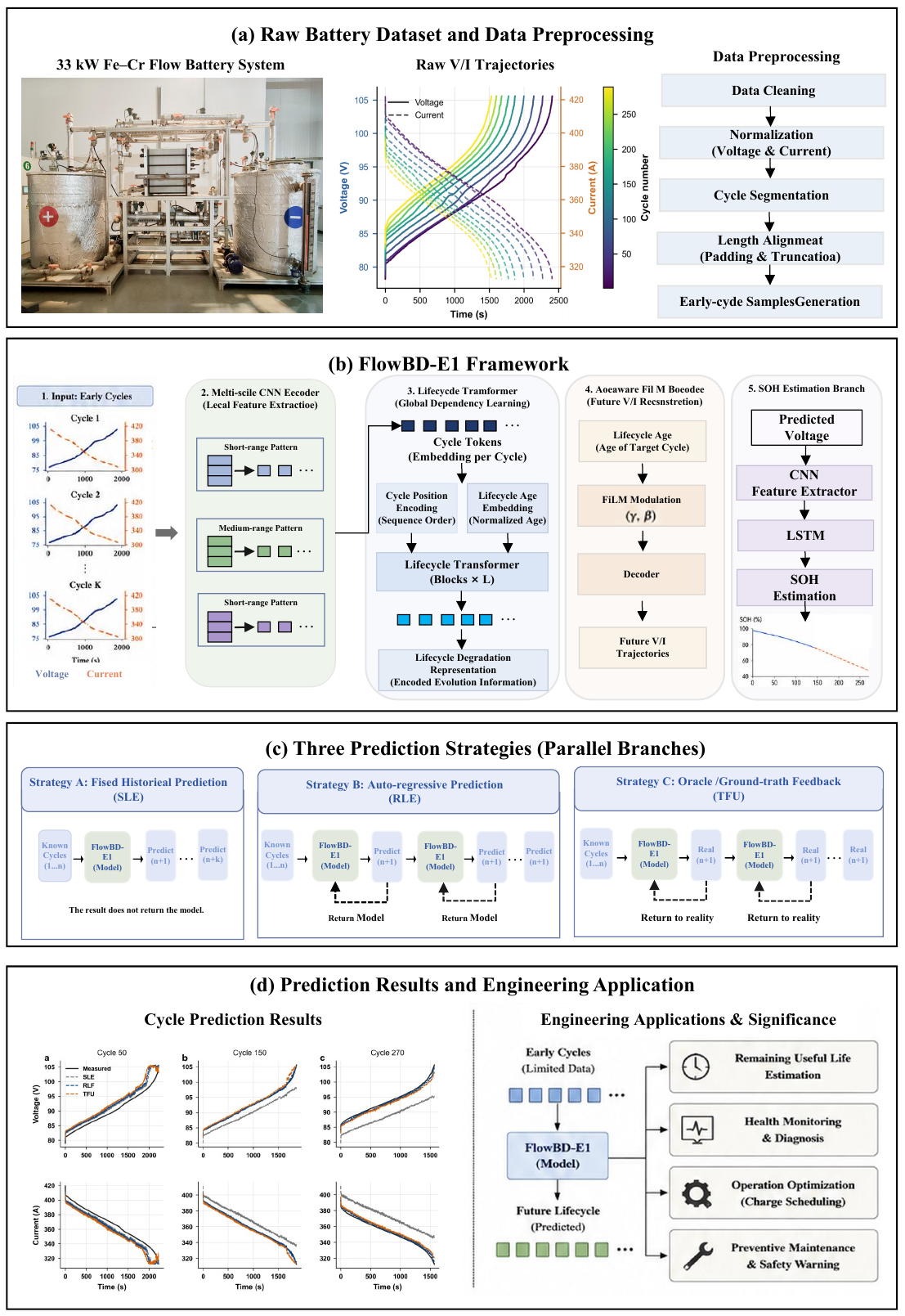}
\caption{Industrial Fe-Cr RFB dataset, preprocessing workflow and FlowBD-E1 architecture.}
\label{fig:1}
\end{figure*}

The raw dataset contains voltage and current trajectories from a 33 kW Fe-Cr flow-battery system. The preprocessing pipeline cleans the data, normalizes voltage and current, segments charge cycles and aligns variable-length curves by padding or truncation. FlowBD-E1 encodes early-cycle V/I curves with a multi-scale CNN encoder, learns lifecycle dependency with a Transformer, reconstructs future trajectories with an age-aware FiLM decoder and derives SOH from generated curves.

FlowBD-E1 contains four functional blocks. First, a multi-scale CNN encoder maps each aligned V/I trajectory into a latent cycle token, with short-, medium- and long-range convolutional pooling used to retain local waveform features and broader phase structure. Second, a lifecycle Transformer models dependencies across the sequence of cycle tokens. The Transformer receives cycle-position information and normalized lifecycle age so that the same local waveform feature can be interpreted differently at different ageing stages. Third, an age-aware FiLM decoder modulates latent features using the target cycle age and reconstructs full future V/I trajectories. Fourth, a downstream health branch estimates SOH either directly from reconstructed voltage features or indirectly from curve integration.

Three prediction strategies were evaluated. SLE uses the early-cycle context to generate one latent state and reuses it for all future cycles with updated age embeddings. RLF predicts one future latent state at a time, appends the prediction to the context and repeats the process autoregressively. TFU supplies the true encoded latent state for each future cycle and therefore acts as a diagnostic upper bound rather than a deployable strategy.

\subsection{Evaluation metrics}\label{evaluation-metrics}

Prediction accuracy was quantified using MAPE, RMSE and the coefficient of determination ($R^2$). For a measured sequence $y_i$, a predicted sequence $\hat{y}_i$, their mean $\bar{y}$, and N evaluated samples, the metrics were defined as:

\begin{align}
\mathrm{MAPE} &= \frac{100\%}{N}\sum_{i=1}^{N}\left|\frac{y_i-\hat{y}_i}{y_i}\right|, \label{eq:mape}\\
\mathrm{RMSE} &= \sqrt{\frac{1}{N}\sum_{i=1}^{N}\left(y_i-\hat{y}_i\right)^2}, \label{eq:rmse}\\
R^2 &= 1-\frac{\sum_{i=1}^{N}\left(y_i-\hat{y}_i\right)^2}{\sum_{i=1}^{N}\left(y_i-\bar{y}\right)^2}. \label{eq:r2}
\end{align}

\section{Results and discussion}\label{results-and-discussion}

\subsection{Early-cycle information is sufficient for long-horizon trajectory forecasting}\label{early-cycle-information-is-sufficient-for-long-horizon-trajectory-forecasting}

To determine how early a useful lifetime forecast can be made, we varied the number of observed commissioning cycles and predicted the remaining lifecycle. The analysis shows that the problem is learnable from surprisingly little data: with only one known cycle, RLF already reached a joint V/I MAPE of 0.634\%, whereas SLE remained near 3.021\%. When the known window increased to 9 cycles, RLF achieved 0.731\% joint V/I MAPE, with 0.735\% voltage MAPE and 0.726\% current MAPE. This indicates that the early charge curve contains a strong stack fingerprint, including impedance, flow distribution and initial electrolyte condition.

\begin{figure*}[!t]
\centering
\includegraphics[width=\textwidth]{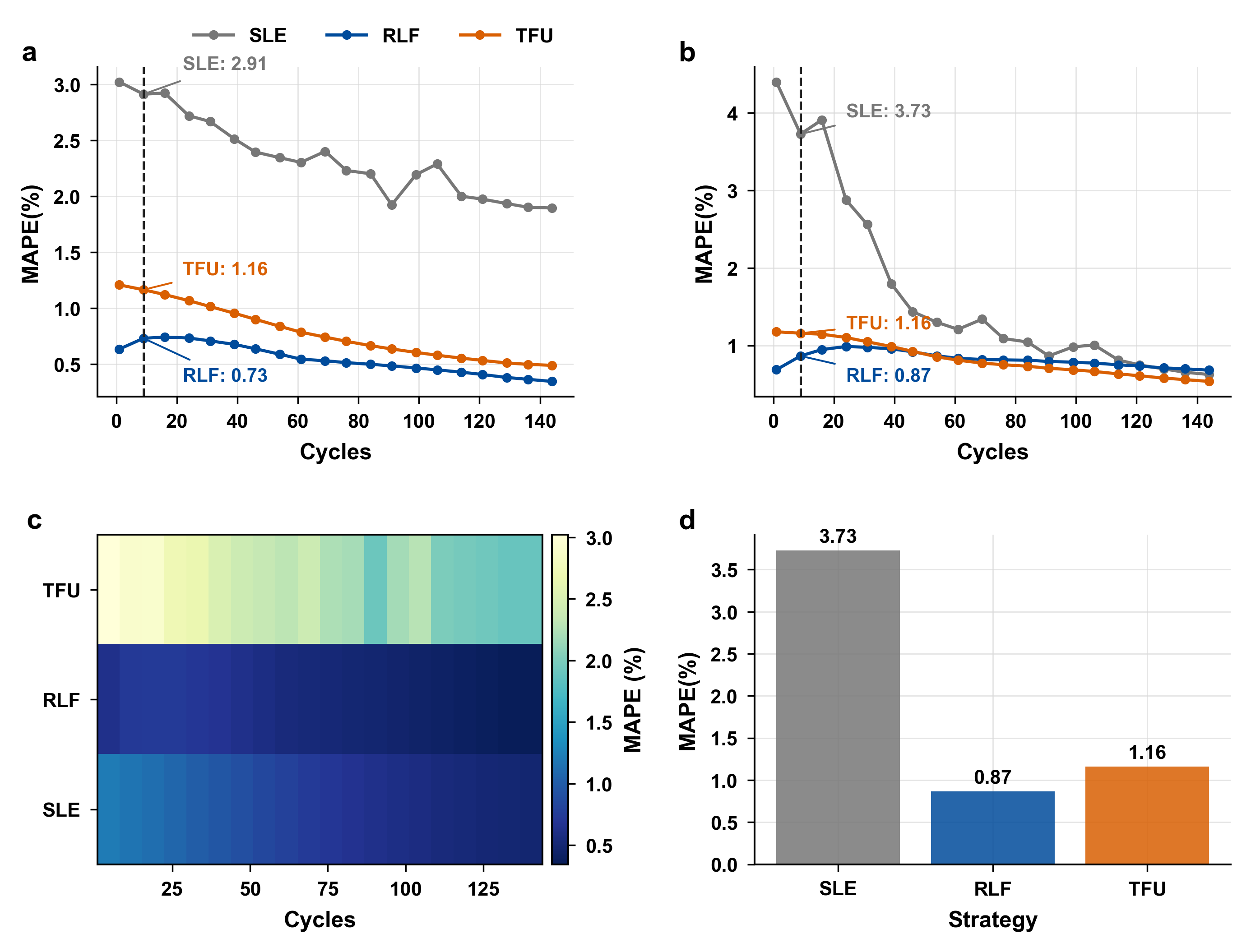}
\caption{Early-cycle feasibility of charge-trajectory prediction.}
\label{fig:2}
\end{figure*}

Figure~\ref{fig:2}a compares joint V/I MAPE as the number of known cycles increases, showing that RLF stays below 1\% across the tested early-window range. Figure~\ref{fig:2}b reports the corresponding SOH MAPE and shows that trajectory generation also supports health estimation. Figure~\ref{fig:2}c visualizes the strategy-by-window error landscape, making the stability of RLF clearer than in the line plots alone. Figure~\ref{fig:2}d summarizes the n = 9 operating point, where RLF provides the lowest deployable error.

Together, these panels reframe early operation as a diagnostic opportunity rather than a waiting period. In practical battery management, operators usually need to run a system for an extended time before degradation trends become visible. FlowBD-E1 instead uses the early waveform to infer a latent degradation path, allowing maintenance and scheduling decisions to be considered during commissioning. The finding also explains why scalar early-cycle descriptors alone may be insufficient: the predictive signal is distributed over the entire curve, not confined to a single voltage, current or capacity value.

\subsection{Recursive latent forecasting gives the best autonomous strategy}\label{recursive-latent-forecasting-gives-the-best-autonomous-strategy}

After identifying n = 9 as a practical early-assessment point, we asked which forecasting strategy can be deployed without future measurements. We therefore compared SLE, RLF and TFU under the same early-window condition. At n = 9, the joint V/I MAPE values were 2.911\% for SLE, 0.731\% for RLF and 1.164\% for TFU in the cross-window summary. The strategy-level diagnostic also showed that RLF reduced error by roughly 70-80\% relative to SLE across most prediction horizons.

\begin{figure*}[!t]
\centering
\includegraphics[width=\textwidth]{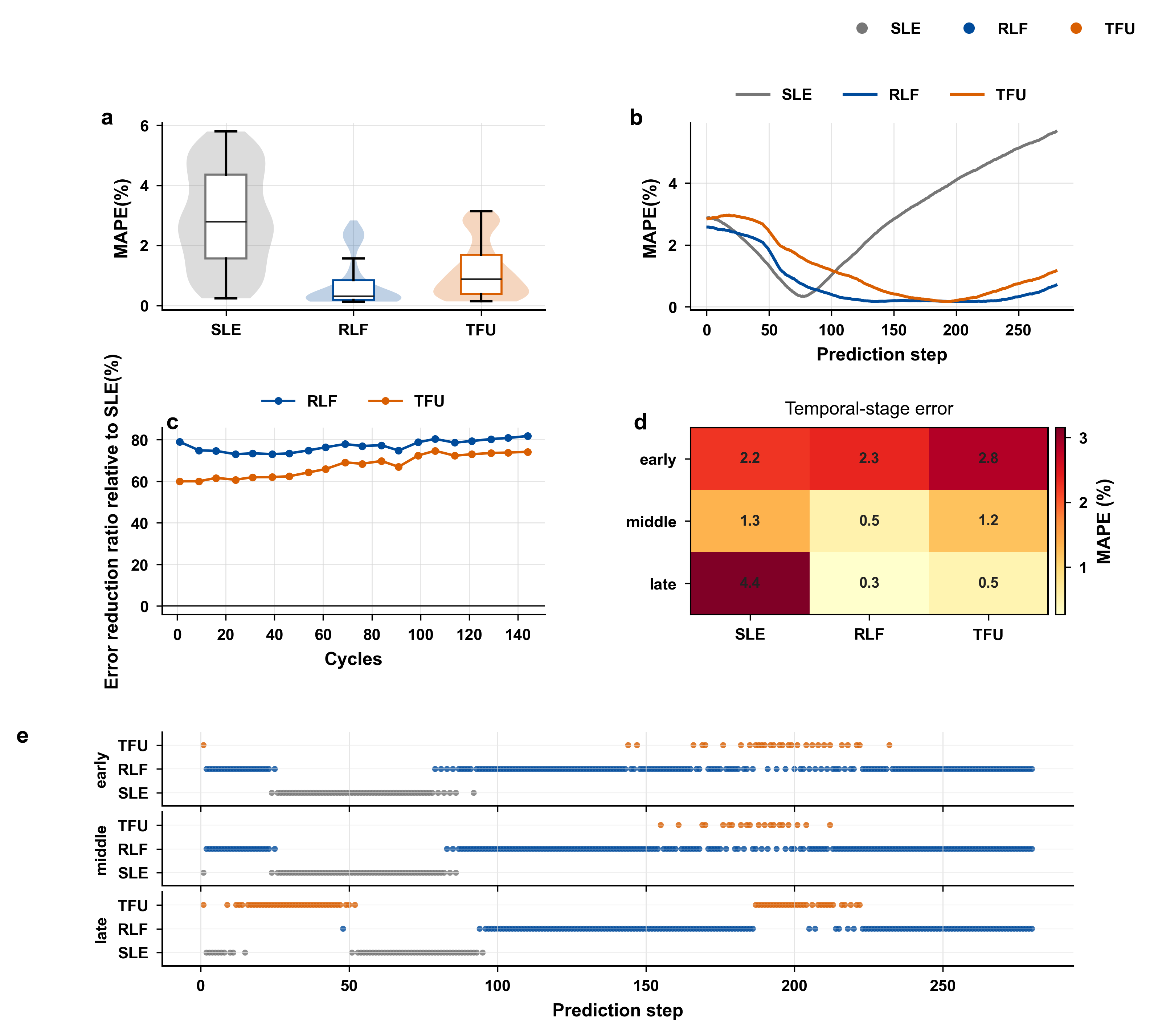}
\caption{Strategy optimization and diagnostic comparison at n = 9.}
\label{fig:3}
\end{figure*}

Figure~\ref{fig:3}a shows the full error distribution and confirms that RLF has the lowest median and narrowest spread among deployable strategies. Figure~\ref{fig:3}b plots error against prediction step, revealing the horizon-dependent accumulation of SLE error and the comparatively stable RLF rollout. Figure~\ref{fig:3}c expresses the error-reduction ratio relative to SLE, showing that RLF preserves most of its advantage throughout the lifecycle. Figure~\ref{fig:3}d separates early, middle and late prediction stages, and Figure~\ref{fig:3}e maps which strategy wins at each prediction step, exposing when TFU acts mainly as a diagnostic reference.

The mechanism behind this difference is visible in the temporal error curves. SLE performs acceptably near the known window but deteriorates as the battery moves further away from the frozen latent representation. RLF avoids this failure mode because each generated latent token becomes part of the evolving context, enabling the model to track gradual ageing in latent space. TFU does not always outperform RLF because access to measured future encodings does not remove decoder error and can expose the model to latent states that are locally accurate but not optimally aligned with the generative rollout. This comparison therefore supports RLF as the deployable strategy while retaining TFU as a useful diagnostic control.

\subsection{Generated trajectories preserve voltage-current physics and SOH evolution}\label{generated-trajectories-preserve-voltage-current-physics-and-soh-evolution}

Low aggregate error is useful only if the generated curves retain the physical structure of charge operation. We therefore inspected representative trajectories and the derived SOH sequence rather than relying only on averaged metrics. RLF closely follows the voltage rise and current decline under constant-power operation. In the early window, it reproduces the full charge duration and the terminal voltage increase; in the middle window, it tracks the shortened curve and shifted voltage profile; and in the late window, it captures the elevated plateau voltage and earlier terminal rise that accompany advanced ageing.

\begin{figure*}[!t]
\centering
\includegraphics[width=\textwidth]{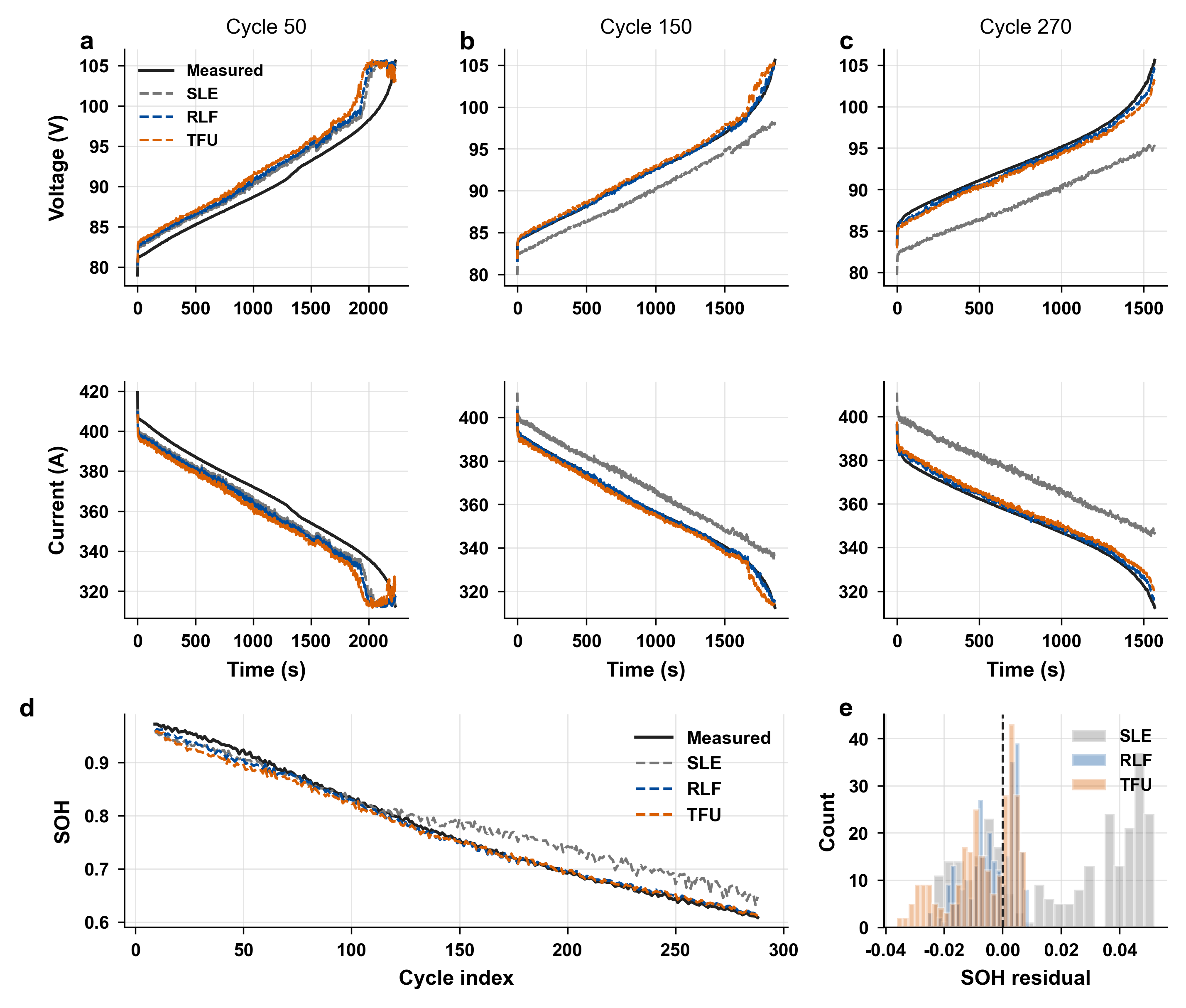}
\caption{Trajectory-level reconstruction and SOH estimation.}
\label{fig:4}
\end{figure*}

\begin{figure*}[!t]
\centering
\includegraphics[width=0.70\textwidth]{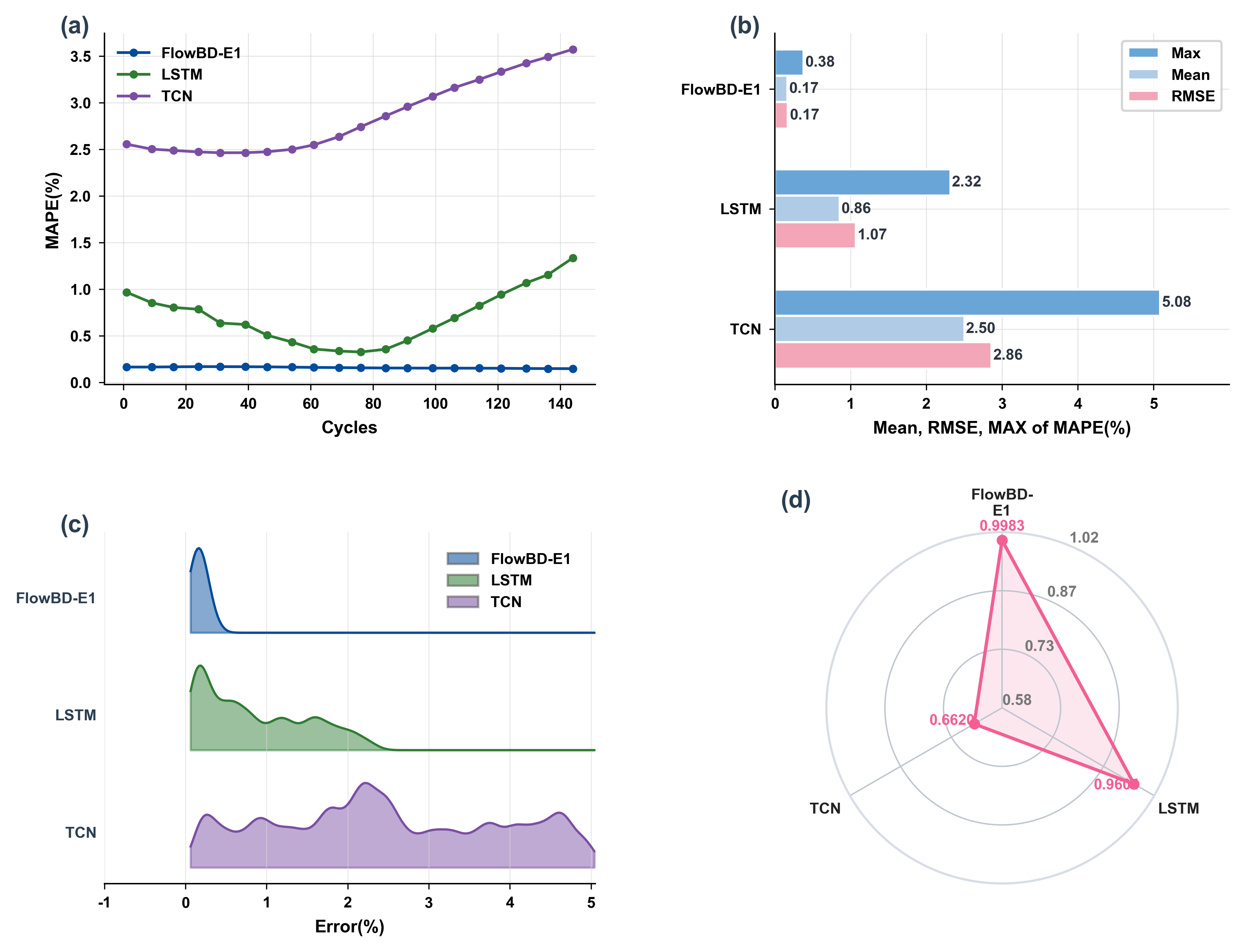}
\caption{Ablation and baseline comparison.}
\label{fig:5}
\end{figure*}

\begin{figure*}[!t]
\centering
\includegraphics[width=0.70\textwidth]{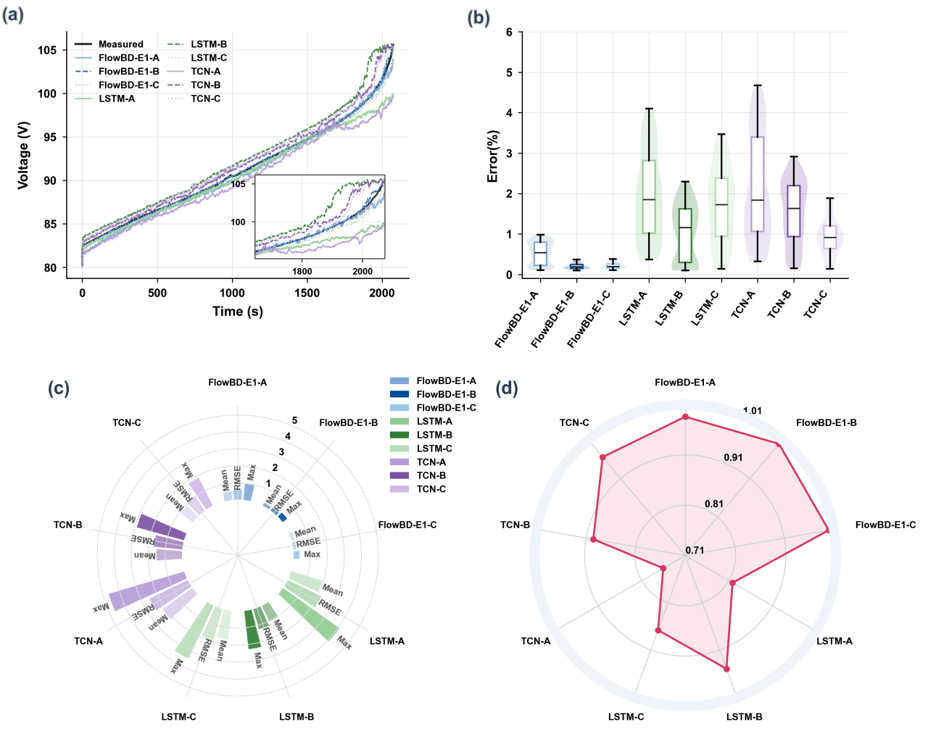}
\caption{Generalization and industrial validation.}
\label{fig:6}
\end{figure*}

Figure~\ref{fig:4}a-c compare measured and predicted voltage/current trajectories at cycles 50, 150 and 270, respectively, demonstrating that the model preserves waveform evolution from early to late life. Figure~\ref{fig:4}d shows that the SOH trajectory derived from generated curves follows the measured decline from approximately 0.96 to 0.61. Figure~\ref{fig:4}e reports the SOH residual distribution, where RLF residuals cluster near zero more tightly than SLE.

This trajectory-level behaviour matters because it preserves the link between prediction and electrochemistry. A scalar SOH forecast can be numerically accurate while hiding whether the model understands the charge process. In contrast, FlowBD-E1 must reproduce two coupled signals: voltage and current. The SOH analysis confirms the value of this generative route. At n = 9, RLF gave a direct SOH MAPE of 0.867\% and an integral-based SOH MAPE of 0.662\%. The integral estimate is especially important because it is derived from the generated V/I curves rather than from a separately trained scalar target. Thus, one generated waveform supports multiple health indicators without retraining the model for each metric.

\subsection{Architecture ablation confirms the role of age-aware generative decoding}\label{architecture-ablation-confirms-the-role-of-age-aware-generative-decoding}

To test whether the proposed architecture contributes beyond a generic sequence model, we compared the full FlowBD-E1 design with LSTM and TCN alternatives. In the n = 9 ablation summary, the full model achieved 0.158\% voltage MAPE and 0.175\% current MAPE under RLF on the 289-cycle sequence, with voltage and current $R^2$ values of 0.9983. The LSTM and TCN variants were less stable, especially under long-horizon rollouts.

Figure~\ref{fig:5}a tracks MAPE across the prediction horizon and shows that FlowBD-E1 remains nearly flat while LSTM and TCN errors grow. Figure~\ref{fig:5}b summarizes maximum, mean and RMSE of MAPE, highlighting the lower average and worst-case errors of FlowBD-E1. Figure~\ref{fig:5}c compares error distributions and shows that FlowBD-E1 concentrates errors near zero. Figure~\ref{fig:5}d gives an aggregate radar-style score, where the full model dominates the baseline variants.

The ablation result clarifies why FlowBD-E1 performs better than generic sequence models. The task is not only to extrapolate a time series; it is to reconstruct future curves whose local features have different meanings depending on lifecycle age. The FiLM decoder supplies this age-dependent modulation, while the Transformer represents long-range cycle-to-cycle dependency. LSTM and TCN baselines can model temporal dependence, but they do not separate lifecycle age, latent degradation state and waveform reconstruction as explicitly. This separation appears to be essential for maintaining low error during the late cycles, where small shifts in capacity and terminal polarization strongly affect curve shape.

\subsection{Industrial validation supports deployment beyond the training sequence}\label{industrial-validation-supports-deployment-beyond-the-training-sequence}

Because deployment requires robustness outside the sequence used for model selection, we finally evaluated the trained framework on an independent industrial validation setting. For n = 9 and a 200-cycle sequence, the full model under RLF achieved 0.208\% voltage MAPE and 0.204\% current MAPE, with $R^2$ values of 0.9976 and 0.9978, respectively. The full model was also the best model across SLE, RLF and TFU in the generalization summary. Compared with the NoFiLM, LSTM and TCN variants, FlowBD-E1 produced lower mean, RMSE and maximum errors, indicating that the model did not merely memorize one sequence.

Figure~\ref{fig:6}a overlays measured and predicted voltage trajectories under the independent validation condition, showing that FlowBD-E1 remains closest to the measured curve. Figure~\ref{fig:6}b compares error distributions across model-strategy combinations and shows that FlowBD-E1 variants occupy the lowest-error region. Figure~\ref{fig:6}c summarizes mean, RMSE and maximum errors in polar form, while Figure~\ref{fig:6}d provides a normalized aggregate score that ranks the full FlowBD-E1 variants above the baselines.

This final result connects the method to practical battery management. Industrial RFB data differ from laboratory cell data because stack operation includes flow, thermal, electrolyte and control-system variability. A model that fails under these conditions would be of limited operational value even if it performed well on curated laboratory datasets. The independent validation suggests that early-cycle trajectory generation is robust enough to support real deployment studies. The present evidence remains bounded: all data come from one Fe-Cr system class and one operating protocol. Future work should therefore test cross-power, cross-temperature, cross-electrolyte and multi-site transfer before claiming chemistry-wide generality.

\FloatBarrier
\section{Conclusions}\label{conclusion}

This study shows that complete future charge trajectories of an industrial Fe-Cr redox flow battery can be forecast from only a small early-cycle record. FlowBD-E1 combines multi-scale waveform encoding, lifecycle Transformer modelling and age-aware FiLM decoding to generate future V/I curves rather than only scalar SOH values. With the first 9 of 289 cycles, RLF achieved 0.731\% joint V/I MAPE and supported SOH estimation below 1\% MAPE. Ablation and independent-sequence validation showed that the full age-aware generative architecture is more stable than LSTM and TCN baselines and retains sub-percent errors in industrial validation. These results support early-cycle generative forecasting as a practical route toward proactive flow-battery management, while also defining the boundary of the present evidence: broader validation across chemistries, operating protocols and field sites is required before the model can be treated as a universal RFB lifetime predictor.

\end{document}